# Belief Maintenance in Bayesian Networks

**Marco Ramoni**
Cognitive Studies in Medicine
McGill Cognitive Science Centre
McGill University, Montreal, Canada

**Alberto Riva**
Laboratorio di Informatica Medica
Dipartimento di Informatica e Sistemistica
Università di Pavia, Pavia, Italy

## Abstract

*Bayesian Belief Networks* (BBNs) are a powerful formalism for reasoning under uncertainty but bear some severe limitations: they require a large amount of information before any reasoning process can start, they have limited contradiction handling capabilities, and their ability to provide explanations for their conclusion is still controversial. There exists a class of reasoning systems, called *Truth Maintenance Systems* (TMSs), which are able to deal with partially specified knowledge, to provide *well-founded* explanation for their conclusions, and to detect and handle contradictions. TMSs incorporating measure of uncertainty are called *Belief Maintenance Systems* (BMSs). This paper describes how a BMS based on probabilitistic logic can be applied to BBNs, thus introducing a new class of BBNs, called *Ignorant Belief Networks*, able to incrementally deal with partially specified conditional dependencies, to provide explanations, and to detect and handle contradictions.

## 1 Introduction

*Bayesian Belief Networks* (BBNs) are a powerful formalism for reasoning under uncertainty. They have been summarized by Pearl (Pearl, 1988), but they have been independently developed by several researchers during the past few years. BBNs have been successfully applied to several domains, from medical diagnosis to natural language understanding.

A BBN is a direct acyclic graph in which nodes represent stochastic variables and arcs represent conditional dependencies among the variables. BBNs are particularly appealing since they are based on a sound probabilistic semantics and they are the reasoning cores of complete decision making systems, called *Influence Diagrams*. From a probabilistic point of view, they provide a straightforward way to represent dependency and independence assumptions among variables, thus making easier the representation and the acquisition of knowledge.

Despite a considerable success, BBNs bear some severe limitations:

**Ignorance:** They require a large amount of information. The number of conditional probabilities needed to specify a conditional dependency grows exponentially with the number of parent variables. This effect leads to serious difficulties since all the conditionals probabilities defining a conditional dependency among variables, as well as all the prior probabilities for the root variables, have to be known before any reasoning process can start. Moreover, each probability assignment expresses in a single number both the belief about an event and the reliability of such a belief. Both these limitations arise from the inability of BBNs to distinguish between *uncertainty* and *ignorance*.

**Explanation:** The ability of BBNs to provide an explanation for their conclusions is still controversial, and it represents a challenge for researchers in the area.

**Consistency:** BBNs improve over the traditional scheme of Bayesian expert systems (Charniak, 1991), since they ensure that if a network is locally consistent, it is also globally consistent. Problems arise when an inconsistent probability value is assigned, because they do not provide any efficient contradiction handling method. It would be better if they were able to identify the set of inconsistent assignments and ask for the retraction of one (or some) of them.

Researchers in Artificial Intelligence have developed a class of reasoning systems, called *Truth Maintenance Systems* (TMSs) (McAllester, 1990), which are able to deal with partially specified knowledge, to provide *well-founded* explanation for their conclusions, and to detect and handle contradictions (Forbus and de Kleer, 1992).

During the past decade, three main classes of TMSs have emerged: *Justification-based* TMSs (JTMSs)

(Doyle, 1979), *Assumption-based* TMSs (ATMSs) (de Kleer, 1986a), and *Logic-based* TMSs (LTMSs) (McAllester, 1980). A JTMS records derivability dependencies among propositions and propagates binary truth-values along chains of dependencies. The ATMS also records derivability dependencies among propositions, but rather than propagating truth-values it labels each proposition with the minimal consistent set of assumptions under which it can be derived. On the other hand, a LTMS manipulates full propositional formulas built from propositions and truth-functional connectives, rather than networks of dependencies, and propagates truth-values using a unit resolution style algorithm called *Boolean Constraint Propagation* (BCP) (McAllester, 1990).

Several attempts have been made to include probabilities in TMSs. TMSs that are able to reason on the basis of probabilistic rather than binary truth-values are called *Belief Maintenance Systems* (BMSs). Falkenhainer (Falkenhainer, 1986) developed a *Justification-based* BMS, introducing Dempster-Shafer (Shafer, 1976) belief functions in a JTMS. De Kleer and Williams (de Kleer and Williams, 1987) augmented the ATMS formalism with probability measures. Laskey and Lehner (Laskey and Lehner, 1989) proved a formal equivalence between the belief of a proposition computed by a Dempster-Shafer function and the probability of the ATMS-label of the proposition, and they provide a correct algorithm for computing the beliefs of these labels. D'Ambrosio (D'Ambrosio, 1987) exploited the ATMS architecture to compute a special case of belief functions.

We have extended the third class of TMSs to probability, thus producing a *Logic-based* BMS(LBMS) (Ramoni and Riva, 1993). This paper will describe how this method for belief maintenance can be applied to BBNs, thus introducing a new class of BBNs able to incrementally deal with partially specified conditional dependencies and prior probabilities, to provide well-founded explanations for their conclusions, and to detect and handle contradictions.

## 2 Logic-based Belief Maintenance

In this section, we will summarize the description of a new kind of BMS, called *logic-based* BMS, in which the Boolean operators of standard logic act as constraints on the probabilistic truth-values of propositions. The LBMS can be regarded as a generalization to interval truth-values of the BCP used by the LTMS. As the LTMS is based on standard propositional calculus, the LBMS is based on the (propositional fragment of) probabilistic logic.

### 2.1 Preliminaries

The LBMS assumes a propositional language defined by an infinite set of *atomic propositions* $S = \{a_1, \ldots, a_n\}$ and by the standard Boolean operators $\neg, \vee, \wedge, \supset$, and $\equiv$. A *literal* is an atomic proposition $a_i$ or its negation $\neg a_i$. An atomic proposition $a_i$ is a *positive* literal and the negation of an atomic proposition $\neg a_i$ is a *negative* literal. A *clause* is a finite disjunction of literals. We will use the term *formula* to refer to a generic legal sentence in the language, and we will denote them with $f_1, f_2, \ldots$. When no ambiguity can arise, we will refer to atomic propositions simply as propositions. The LBMS knows also a distinguished subset of propositions $\mathcal{A} \subset \mathcal{S}$ called *assumptions*. An assumption is a literal initially set as having a particular truth-value.

We now need an evaluation function over the formulas of our language. We first introduce an additive real-valued function $P_0(f_i) = p$ satisfying the axioms of conventional probability calculus. Unfortunately, Boolean operators are unable to constrain propositions to point-valued probability. Given $P_0(f_1 \vee f_2) = p$, $P_0(f_1)$ and $P_0(f_2)$ can range between 0 and $p$. Hence, the evaluation function for our language has to be a function $P(f_i) = [p_*, p^*]$ such that for any formula $f_i$ of our language, $p_* \leq p \leq p^*$. The function $P(f_i)$ assigns to $f_i$ an interval truth-value: we will denote with $P_*(f_i) = p_*$ and $P^*(f_i) = p^*$ the lower and the upper bounds of the function $P(f_i) = [p_*, p^*]$, respectively.

### 2.2 Probabilistic Logic

Probabilistic logic (Nilsson, 1986) provides a semantic framework for extending the standard (Boolean) concept of satisfaction to a probabilistic one, that can be interpreted in terms of the Venn Diagram representation of the probability of a proposition. The probability $P_0(a_i)$ of a proposition $a_i$ is bounded by the following inequality:

$$P_0(a_j) + P_0(a_j \supset a_i) - 1 \leq P_0(a_i) \leq P_0(a_j \supset a_i) \quad (1)$$

Inequality (1) may be regarded as the probabilistic interpretation of *modus ponens*: from $P_0(a_j) = p_1$ and $P_0(a_j \supset a_i) = p_2$ we can derive bounds of probability representing the truth-value of $a_i$. Moreover, it is a special case of a more general inequality that applies to any clause. Let $C = \bigvee_{i=1}^{n} a_i$ be a clause, the probability of $a_i$ is bounded by the following inequality:

$$P_0(C) - \sum_{j \neq i} P_0(a_j) \leq P_0(a_i) \leq P_0(C) \quad (2)$$

The right hand side of (2) is obvious: no proposition may have a probability greater than the maximum probability of any disjunction it is part of. In set-theoretic terms, this means that a set cannot be larger than its union with other sets. The left hand side states that the minimum probability of a proposition has to be equal to or greater than the difference between the probability of any clause $C$ in which it appears and the probability of any literal appearing in $C$. Unfortunately, the constraints directly derived from

500 Ramoni and Riva

inequality (2) turn out to be too weak: the bounds they produce are too wide, thus including inconsistent values. The INFERNO system (Quinlan, 1983), that is usually regarded as a local approximation to probabilistic logic (Pearl, 1988; Nilsson, 1993), exhibits this kind of behavior, in producing wider bounds. Because of that, INFERNO has been strongly criticized (Pearl, 1988).

## 2.3 Constraints

The weakness of the constraints derived from (2) arises from too strong an enforcement of their *locality* based on the assumption that all proposition in a clause are pairwise disjoint and, in the Venn Diagram representation of a clause, the intersection of all propositions is always empty. It is apparent that this assumption is too strong.

In order to drop this assumption, we need to represent this intersection among propositions in a clause. We call it *overlapping factor*. The overlapping factor of a clause $C = \bigvee_{i=1}^{n} a_i$ is defined as

$$\sum_{\lambda_1=0}^{1} \cdots \sum_{\lambda_n=0}^{1} P(\bigwedge_{i=1}^{n} a_i^{\lambda_i}) \cdot \Delta(\lambda_1, \ldots, \lambda_n) \quad (3)$$

where $\{a_1, \ldots, a_n\}$ are atomic propositions, $a^1 = a$, $a^0 = \neg a$, and the function $\Delta$ is defined as:

$$\Delta(\lambda_1, \ldots, \lambda_n) = \max\{0, (\sum_{i=1}^{n} \lambda_i) - 1\}$$

In order to compute the overlapping factor of a clause $C$, we need to know the probability of those clauses that contain exactly the same set of atomic propositions, and hence to abandon the strict locality of clauses.

Generalizing definition (2) to interval truth-values, we derived a set of constraints on the minimum and maximum probability of propositions (Ramoni and Riva, 1993), by dropping the assumption that all literals in a clause are pairwise disjoint.

The first constraint enforces the left hand side of (2).

**Constraint 1** *The probability of a proposition $a_i$ in clause $C$ is bounded by:*

$$P_*(a_i) \geq P(C) + \mathcal{F}_C - \sum_{j \neq i} P^*(a_j)$$

*where $\mathcal{F}_C$ is the overlapping factor of clause $C$.*

The second constraint is directly derived from the well known *Additivity* axiom which states that if $a_i$ is an atomic proposition, and $\{\phi_1, \ldots, \phi_{2^n}\}$ is the set of all the conjunctions that contain all possible combinations of the same $n$ atomic propositions negated and unnegated, then:

$$P(a_i) = \sum_{j=1}^{2^n} P(a_i \wedge \phi_j) \quad (4)$$

**Constraint 2** *The minimum probability of a proposition $a$ is bounded by:*

$$P_*(a) \geq 1 - \sum_{i=1}^{2^n} (1 - P(\neg a \vee \psi_i))$$

In constraint 2, we replaced the equality in (4) with an inequality because the constraint 2 holds also when only a subset of the clauses $\{(\neg a_i \vee \psi_1), \ldots, (\neg a_i \vee \psi_{2^n})\}$ is known, and causes $P_*(a_i)$ to increase monotonically as the number of known clauses increases. Hence, if $\neg(\neg a_i \vee \psi_j)$ is the clause obtained by the application of De Morgan's laws, we have $P^*(a_i \wedge \phi_j) = 1 - P_*(\neg a_i \vee \psi_j)$. It is worth noting that the constraint 2 subsumes the right hand side of inequality (2).

From the definitions above, we can easily derive a definition of *inconsistency* in the LBMS network. An inconsistency can arise when, for any proposition $a_i$ in the network:

$$P_*(a_i) > P^*(a_i) \quad (5)$$

or when, for a clause $C$ in the network:

$$\sum_{i=1}^{n} P^*(a_i) - \mathcal{F}_C < P_*(C) \quad (6)$$

Where $\mathcal{F}_C$ is the overlapping factor of the clause $C$. If $P_*(a_i) > P^*(a_i)$, then no probability function $P_0(a_i)$ can exist such that $P_*(a_i) \leq P_0(a_i) \leq P^*(a_i)$. When condition (6) is met, there is no way to satisfy $C$ since the sum of all maximum probabilities of propositions does not cover the minimum probability of $C$.

## 2.4 Propagation

In a LTMS, each clause represents a logical constraint on the truth-values of its propositions. To be satisfied, a clause must contain at least one proposition whose truth-value is consistent with its sign in the clause: *true* if the proposition appears unnegated in the clause, *false* if it appears negated. When all propositions but one violate their sign in a clause, the clause is said to be *unit-open*, and the LTMS forces the remaining proposition to have the truth-value indicated by its sign, thus preventing the assignment of inconsistent truth-values. Contradiction may arise in two ways: a proposition is labeled both *true* and *false* or a clause is completely *violated*, i.e. each of its propositions has a truth-value opposite to its sign in the clause. In the LBMS, these two situations correspond to the inconsistency conditions (5) and (6), respectively.

In the LBMS, we have two constraints to apply. The constraint 2 is applied only when a new clause is added to the network. The application of this constraint exploits its incremental character. The current implementation uses a set of *tables* each of which stores all the clauses containing the same set of propositions and the constraint is applied to each literal in the clause. Moreover, this method allows us to incrementally record the overlapping factor of the clauses currently known by the LBMS. Constraint 1 is applied when the maximum probability of all the literals in the clause $C$ but one is less than the probability of $C$. In this case, the clause is *unit open*, and the constraint increases appropriately the minimum probability of the remaining literal. The algorithm for applying the constraint is basically a Waltz's propagation algorithm extended to intervals, and it is described in (Ramoni and Riva, 1993): each proposition is labeled with a set of possible values, and the constraints (in our case, the application of the above defined constraints to the clauses) are used to restrict this set. The LBMS can exhibit this behavior because if a clause is satisfied for a given truth-value of a proposition $P(a_i) = [p_*\ p^*]$, it will be satisfied for any subset of $[p_*\ p^*]$. This property, which is implicit in the form of the inequalities in our constraints, implies a monotonic narrowing of the truth-values, thus ensuring the incrementality of the LBMS.

### 2.5 Properties

Extending the usual logical concepts of *soundness* and *completeness* from Boolean values to probability intervals (Grosof, 1986), we can say that the system defined by the constraints 1 and 2 is probabilistically *sound* (i.e. it returns intervals that are equal to or wider than the intended ones), but it is not *complete* (i.e. it does not return intervals that are equal to or stricter than the intended ones). This incompleteness is due to the fact that the LBMS calculates the overlapping factor of a clause $C$ using just a particular set of clauses (i.e., those that contain exactly the same set of atomic propositions as $C$) and does not exploit the other sets of clauses that define the overlapping factor of $C$, for example, the powerset of all propositions contained in clause $C$.

There are two motivating factors behind the choice of this particular set of clauses. First of all, we found that the calculation of the overlapping factor is the only source of complexity in the LBMS which, being assimilable to the BCP, runs in linear time and space with respect to the number of clauses. Since we have devised an efficient method to calculate the overlapping factor and to apply constraint 2, and since probabilistic entailment is known to be intractable in nature (Nilsson, 1993), the incompleteness of the LBMS represents a compromise between functionality and efficiency. Furthermore, the representation in the LBMS of a probabilistic model expressed in terms of *conditional probabilities* produces a set of clauses that is exactly the one needed to calculate the overlapping factor and to apply the constraint 2. The representation of conditional probabilities in the LBMS is straightforward using the Chain Rule:

$$P_0(a_2|a_1) \cdot P_0(a_1) = P_0(a_1 \wedge a_2) \qquad (7)$$

The resulting conjunction is converted into clausal form through De Morgan's laws and it is then communicated to the LBMS. For instance, the probabilistic model defined by the two conditionals $P_0(a_2|a_1) = 0.2$ and $P_0(a_2|\neg a_1) = 0.6$ with $P_0(a_1) = 0.5$ may be expressed by the set of clauses: $P_0(a_1 \vee a_2) = 0.8, P_0(a_1 \vee \neg a_2) = 0.7, P_0(\neg a_1 \vee a_2) = 0.6, P_0(\neg a_1 \vee \neg a_2) = 0.9$.

## 3 Ignorant Belief Networks

Using the LBMS, it is possible to develop a new class of BBNs based on the LBMS and henceforth able to reason with partially specified conditional dependencies (i.e. lacking some conditional probabilities) and interval probability values. We call these BBNs *Ignorant Belief Networks* (IBNs).

### 3.1 Definitions

A BBN is a direct acyclic graph in which nodes represent stochastic *variables* and arcs represent conditional *dependencies* among those variables. A variable is defined by a finite set of *states* representing the assignment of a value to the variable. Each state in a variable is evaluated by a probability value. Each dependency is defined by the set of all *conditional* probabilities relating the states of the parent variables to the states of the children variables. We will now describe how these definitions can be translated into the LBMS network.

**Variables** In a BBN, all the states of a variable are mutually exclusive and exhaustive: the probability values assigned to all the states in a variable have to sum to unit. In an IBN, when a variable is defined, each state is communicated to the LBMS as a proposition. Moreover, a set of clauses is installed to ensure that the states of the variable are mutually exclusive and exhaustive. For all propositions $a_1, \ldots, a_n$ in the LBMS representing the states of the variable, the disjunction $a_1 \vee \cdots \vee a_n$ and all the conjunctions $\neg(a_i \wedge a_j)$ (with $i \neq j$) are asserted as true in the LBMS. When a probability value is assigned to a proposition $a_i$ representing a state of the variable, the LBMS receives the clause $P^*(a_{i+1} \vee \ldots \vee a_n) = \sigma$, where $\{a_{i+1}, \ldots, a_n\}$ is the set of proposition representing those states in the variable that are still *unknown*, and $\sigma = 1 - \sum_{k=1}^{i} P_*(a_k)$, i.e. the sum of the minimum probabilities of all *known* states in the variable.

**Dependencies** Conditional dependencies among variables are defined by the conditional probabilities among all the states of each variable. In an IBN, a conditional $P(a_i|a_1,\ldots,a_k) = [p_*\ p^*]$, with $i \notin \{1,\ldots,k\}$, is represented as a consumer attached to each proposition $a_1 \ldots a_k$. When the probability value of all states represented by the propositions $a_1 \ldots a_k$ is assigned, the two different clauses resulting from the application of the De Morgan's laws to $(a_i \wedge a_1 \wedge \ldots \wedge a_k)$ and $(\neg a_i \wedge a_1 \wedge \ldots \wedge a_k)$ are communicated to the LBMS. $P(a_i \wedge a_1 \wedge \ldots \wedge a_k)$ and $P(\neg a_i \wedge a_1 \wedge \ldots \wedge a_k)$ are calculated according to a version of the Chain Rule extended to intervals:

$$P_*(a_i \wedge a_1 \wedge \ldots \wedge a_k) = \prod_{j=1}^{k}(P_*(a_j))P_*(a_i|a_1,\ldots,a_k)$$

$$P^*(a_i \wedge a_1 \wedge \ldots \wedge a_k) = \prod_{j=1}^{k}(P^*(a_j))P^*(a_i|a_1,\ldots,a_k)$$

$$P_*(\neg a_i \wedge a_1 \wedge \ldots \wedge a_k) = \prod_{j=1}^{k}(P_*(a_j))(1 - P^*(a_i|a_1,\ldots,a_k))$$

$$P^*(\neg a_i \wedge a_1 \wedge \ldots \wedge a_k) = \prod_{j=1}^{k}(P^*(a_j)) \cdot (1 - P_*(a_i|a_1,\ldots,a_k))$$

The direction of a conditional dependency can be reversed by using the *Inversion Rule* and applying the above defined constraints to the resulting conditionals.

### 3.2 Propagation

From the theory of the TMSs, the LBMS inherits the concept of *consumer* (de Kleer, 1986b). A consumer is a forward-chained procedure attached to each proposition, that is fired when the truth-value of the proposition is changed. The BMSs theory extends the definition of consumers from Boolean to probabilistic truth-values. In the LBMS, a consumer can be defined as fireable when the minimum probability of its proposition is raised, the maximum probability is lowered, or when the difference between the maximum and minimum probability is decreased.

When a variable is defined in the IBN, for each proposition representing its states in the LBMS, two different consumers are defined. The first consumer is used to communicate to the LBMS the clause $P^*(a_{i+1} \vee \ldots \vee a_n) = \sigma$ above defined, in order to enforce the exhaustivity and exclusivity among states in a variable. A second consumer is used to encode the *conditional probabilities* among states, and it is defined when a conditional dependency is installed in the network. When it is fired, it applies the Chain Rule to the defined conditional and communicates the appropriate clauses to the LBMS. A prior probability assignment to a state in a variable is communicated to the LBMS by *assuming* the corresponding proposition with the assigned probability. When the proposition is assumed, the attached consumers are fired, thus starting the propagation process.

Using consumers, IBNs do not perform any computation themselves, but rather act as a high-level knowledge representation language, while the propagation of probabilities is performed by the LBMS. Hence, the computational cost of a propagation grows linearly in space and time with respect to the number of conditional probabilities, even if the number of conditional probabilities needed to specify a conditional dependency grows exponentially with the number of parent variables in the dependency.

### 3.3 Properties

In the introduction, we claimed that the use of a LBMS to develop BBNs could enhance their ability to deal with partially specified probabilistic information, to provide explanations and to handle contradictions.

**Ignorance**   We can identify two different kinds of ignorance that can be represented in this framework: complete ignorance about a conditional probability and partial information about a conditional or prior probability in the network. Since the probability of both propositions and clauses in the LBMS is represented by probability intervals, IBNs are endowed with the ability to express both interval conditional probabilities and interval prior probabilities. Moreover, since conditionals are locally defined and propagated, the reasoning process can start even without the full definition of the joint probability distribution. These features enable the IBNs to represent both the complete ignorance of a conditional probability and the partial information about a conditional or a prior probability.

**Explanation**   TMSs provide rational explanation for their conclusions by describing how these conclusions follow from the current set of assumptions (Forbus and de Kleer, 1992): they not only trace back the set of assumptions responsible for the conclusion but they also describe the derivation steps that lead from those assumptions to the conclusion to be explained.

In the LBMS, each proposition is labeled with an interval truth-value. Hence, the LBMS has to explain the assignment of both its lower and upper bound, that could have been derived from different assumptions trough different paths. In the current implementation, the lower and the upper bounds of the interval associated to each propositions are indexed by the clause that set them during the constraint propagation. To explain the assignment of a minimum or of a maximum probability to a proposition, the LBMS uses a quite simple algorithm. If no clause is supporting the assignment, then the proposition is itself the assumption responsible for its own assignment. If the clause sets the value trough the application of the constraint 2, no assumption is responsible for the assignment because it is directly set by the clause. If the assignment is due to the application of the constraint 1, all the the literals but the one to be explained had their maximum probability lowered and the LBMS tries to explain this lowering. When an assumption is reached, it is returned as an explanation, together with the path followed to reach it. In the IBN, this mechanism has to be enhanced since the assumption of a proposition also causes the generation of clauses, namely, those clauses generated by conditionals and those supplementary clauses needed to enforce the exhaustivity and exclusivity among states in a variable. These clauses are labeled with the proposition that generated them, so that the explanatory algorithm of the IBN is able to follow also these additional paths in order to reach the assumptions and return them as explanations.

| Priors | Step 1 | Step 2 | Step 3 |
|---|---|---|---|
| [fire:yes] | [0.7 0.9] | [0.8 0.8] | [0.8 0.8] |
| [fire:no] | [0.1 0.3] | [0.2 0.2] | [0.2 0.2] |
| [tampering:yes] | [0.85 0.95] | [0.9 0.9] | [0.9 0.9] |
| [tampering:no] | [0.05 0.15] | [0.1 0.1] | [0.1 0.1] |
| **Conditionals** | | | |
| [smoke:yes] \| [fire:yes] | [0.9 0.9] | [0.9 0.9] | [0.9 0.9] |
| [smoke:yes] \| [fire:no] | — | — | [0.01 0.01] |
| [alarm:yes] \| [fire:yes] ∧ [tampering:yes] | [0.5 0.5] | [0.5 0.5] | [0.5 0.5] |
| [alarm:yes] \| [fire:yes] ∧ [tampering:no] | [0.99 0.99] | [0.99 0.99] | [0.99 0.99] |
| [alarm:yes] \| [fire:no] ∧ [tampering:yes] | — | — | [0.85 0.85] |
| [alarm:yes] \| [fire:no] ∧ [tampering:no] | [0.0001 0.0001] | [0.0001 0.0001] | [0.0001 0.0001] |
| [leaving:yes] \| [alarm:yes] | [0.8 0.9] | [0.8 0.9] | [0.88 0.88] |
| [leaving:yes] \| [alarm:no] | [0.001 0.001] | [0.001 0.001] | [0.001 0.001] |
| [report:yes] \| [leaving:yes] | [0.7 0.8] | [0.7 0.8] | [0.75 0.75] |
| [report:yes] \| [leaving:no] | [0.01 0.01] | [0.01 0.01] | [0.01 0.01] |

Table 1: Prior and conditional probabilities defining the network of the example.

**Consistency** One of the fundamental tasks of TMSs is to detect and recover inconsistencies during the problem solving process. Since a BBN is known to be globally consistent if it is locally consistent, the LBMS maintains the consistency of a IBN by checking if one of the inconsistency conditions (5) and (6) is met. When an inconsistency arises, the LBMS uses the same algorithm exploited in the explanatory process to trace back the assumptions responsible for the inconsistency, and ask for their retraction. The algorithm for retraction has been extended in order to take into account that the assumption of a state in a IBN generates additional clauses. Hence, in the current implementation, when a state in the IBN is retracted, the clauses generated by its assumption are deleted, and the constraints they impose over the network are relaxed.

### 3.4 An Example

In a standard BBN, all the conditional probabilities that make up a conditional probability distribution are needed before any reasoning process can start. We will now show with an example how the IBN is instead able to reason from incomplete conditional probability distributions and underspecified prior probabilities. The example is composed of three steps. In the first step, we will communicate to the system the conditional probabilities listed in the first column of Table 1, and we will assign interval prior probabilities to the states in the root variables. Note that two conditional probabilities are missing, and two have an interval value. In the second step, we will refine the prior probabilities turning them into point-valued probabilities, and in the third step we will use the complete conditional probability distribution. Figures 1-4 show the graphical representation of the networks generated in the various steps of the example. The pop-up windows over the variables graphically describe the probability interval (subset of [0 1]) associated to each one of their states. In each bar, the area between 0 and $P_*(a_i)$ is black, the area between $P^*(a_i)$ and 1 is white, and the area between $P_*(a_i)$ and $P^*(a_i)$ is gray. Thus, the width of the gray area is proportional to the ignorance about the probability.

In Step 1, we have communicated to the IBN the conditional probabilities listed in the first column of Table 1. The two conditional probabilities $P([smoke:yes] \mid [fire:no])$ and $P([alarm:yes]|[fire:no] \wedge [tampering:yes])$ are missing, and the two conditional probabilities $P([leaving:yes] \mid [alarm:yes])$ and $P([report:yes] \mid [leaving:yes])$ have interval values. Moreover, we have assumed the interval [0.7 0.9] as the prior probability for [fire:yes] and the interval [0.85 0.95] as the prior probability for [tampering:yes]. All the other probability intervals appearing in Figure 1 were set by the propagation algorithm.

Figure 2 shows a portion of the LBMS network generated by the propagation. Rectangles represent propositions and ovals are clauses. A solid arc linking a proposition to a clause means that the proposition appears unnegated in the clause, while a dashed arc means that it appears negated. The side bars display the minimum and maximum probability values. The thicker border of the proposition [fire:yes] indicates that it is an assumption.

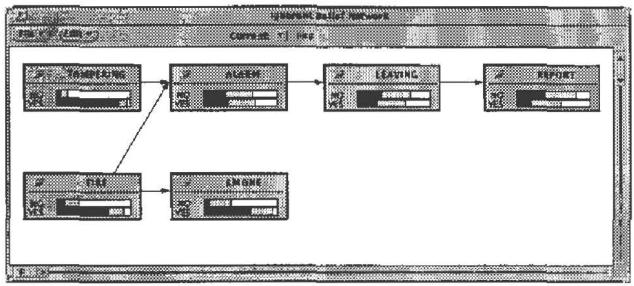

**Figure 1:** The IBN defined by the underspecified conditional model in Table 1 and interval prior probabilities.

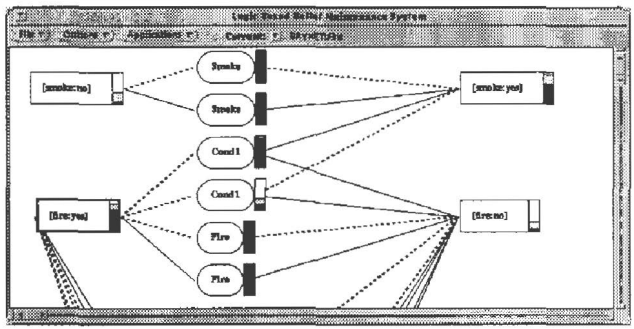

**Figure 2:** A section of the LBMS network defined by the propagation of consumers for the IBN in Figure 1.

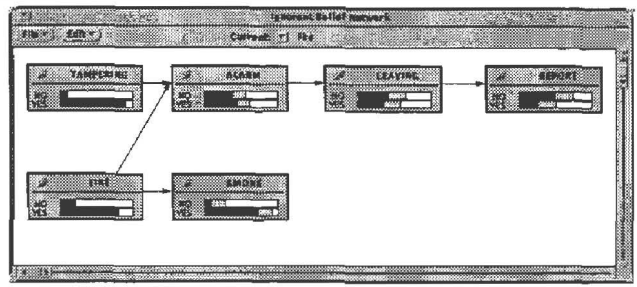

**Figure 3:** The IBN defined by the underspecified conditional distribution in Table 1 and point-valued prior probabilities.

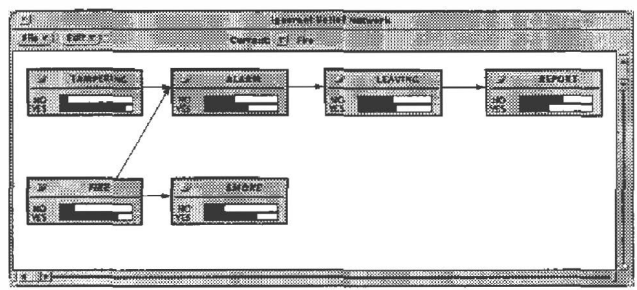

**Figure 4:** The IBN defined by the complete conditional distribution in Table 1 and point-valued prior probabilities.

The clauses $P([\text{fire:yes}] \vee [\text{fire:no}]) = [1\ 1]$ and $P(\neg[\text{fire:yes}] \vee \neg[\text{fire:no}]) = [1\ 1]$ and the clauses $P([\text{smoke:yes}] \vee [\text{smoke:no}]) = [1\ 1]$ and $P(\neg[\text{smoke:yes}] \vee \neg[\text{smoke:no}]) = [1\ 1]$ enforce the exhaustivity and exclusivity property of the states of the variables Fire and Smoke, respectively. The clauses $P(\neg[\text{fire:yes}] \vee [\text{fire:no}] \vee [\text{smoke:yes}]) = [0.91\ 0.93]$ and $P(\neg\ [\text{fire:yes}] \vee [\text{fire:no}] \vee \neg[\text{smoke:yes}]) = [0.19\ 0.37]$ named Cond1 are generated by the application of the Chain Rule with $P([\text{fire:yes}]) = [0.7\ 0.9]$ and $P([\text{smoke:yes}] \mid [\text{fire:yes}]) = [0.9\ 0.9]$. The underspecified probability of the proposition [smoke:yes] is due both to the interval-valued probability of the proposition [fire:yes] and to the absence of the conditional $P([\text{smoke:yes}] \mid [\text{fire:no}])$.

Figure 3 shows how the probability intervals are updated when we assume point-valued prior probabilities for the states of the root variables in the Step 2. Note that, due to the monotonic and incremental character of the propagation algorithm, all the intervals in Figure 3 are subsets of the intervals in Figure 1.

Finally, in Step 3 the complete conditional probability distribution listed in the third column of Table 1 is used. In this case, all the intervals degenerate to point-valued probabilities, and the IBN converges to the values of a standard BBN, as shown in Figure 4. Table 2 summarizes the probability intervals associated to each proposition in each one of the steps described above.

## 4 Conclusions

We have introduced a new method to deal with partially specified probabilistic models and we have applied it to develop a new class of BBNs, based on a LBMS, able to reason with incomplete information on the basis of an explicit representation of ignorance. Furthermore, the LBMS provides the IBN with the ability of detecting and handling contradictions, and of producing well-founded explanations for its conclusions. We have applied the IBNs to the forecasting blood glucose concentration in insulin-dependent diabetic patients using underspecified probabilistic models directly derived from a database containing the daily follow-up of 70 insulin-dependent diabetic patients, in which a very small subset of the complete conditional model needed to define a BBN was available. Instead of the 19200 conditional probabilities required, only 2262 were available (that is, less than 12%), and most of them were affected by ignorance (the mean difference between the maximum and min-imum probability of the conditionals was 0.19) (Ramoni et al., 1994). Still, the system was able to reason and make predictions, taking into account the ignorance about the distributions.

| Proposition | Step 1 | Step 2 | Step 3 |
|---|---|---|---|
| [fire:yes] | [0.7 0.9] | [0.8 0.8] | [0.8 0.8] |
| [fire:no] | [0.1 0.3] | [0.2 0.2] | [0.2 0.2] |
| [tampering:yes] | [0.85 0.95] | [0.9 0.9] | [0.9 0.9] |
| [tampering:no] | [0.05 0.15] | [0.1 0.1] | [0.1 0.1] |
| [smoke:yes] | [0.63 0.93] | [0.72 0.92] | [0.722 0.722] |
| [smoke:no] | [0.07 0.37] | [0.08 0.28] | [0.278 0.278] |
| [alarm:yes] | [0.332 0.697] | [0.439 0.619] | [0.592 0.592] |
| [alarm:no] | [0.303 0.668] | [0.381 0.561] | [0.408 0.408] |
| [leaving:yes] | [0.266 0.664] | [0.352 0.576] | [0.522 0.522] |
| [leaving:no] | [0.336 0.734] | [0.424 0.648] | [0.478 0.478] |
| [report:yes] | [0.19 0.614] | [0.25 0.51] | [0.396 0.396] |
| [report:no] | [0.386 0.81] | [0.49 0.75] | [0.604 0.604] |

Table 2: The probability intervals associated to propositions in the successive steps of the example.

Since BBNs are the reasoning cores of general decision making systems, called *Influence Diagrams*, we are extending our work to develop a new class of Influence Diagrams (a sort of *Ignorant Influence Diagrams*) able to deal with sets of admissible decisions given bounds on distributions over the expected utilities.

### Acknowledgments

This research was supported in part by a grant from the Social Science and Humanities Council of Canada (410 92 1535), and by the AIM Programme of the Commission of the European Communities (A2034). The authors thank Riccardo Bellazzi, Mario Stefanelli, and Tony Marley for their helpful suggestions and discussions.